%% file: main.tex
\documentclass[10pt,twocolumn,letterpaper]{article}

\usepackage[pagenumbers]{cvpr} % To force page numbers, e.g. for an arXiv version
\usepackage{microtype}
\input{preamble}
\definecolor{cvprblue}{rgb}{0.21,0.49,0.74}
\usepackage[pagebackref,breaklinks,colorlinks,allcolors=cvprblue]{hyperref}
\usepackage{multirow}
\usepackage{booktabs,tabularx}
\def\confName{CVPR}
\def\confYear{2027}

\title{POINT: Robust Humanoid Tennis on Real-World Courts\\ with Single-View Perception}

\author{
  Yiming Ren$^{1,*}$ \quad Yichuan Yu$^{1,*}$ \quad Aoru Xue$^{1,*}$ \quad
  Yujing Sun$^{2,*}$ \quad Zhenxuan Wu$^{1}$ \\
  Gang Wang$^{3}$ \quad Wenhan Liu$^{3}$ \quad Feixiang Lu$^{4}$ \quad
  Zhouye Chen$^{3}$ \quad Yuexin Ma$^{1,\dagger}$ \\[4pt]
  {\normalsize $^{1}$ShanghaiTech University \qquad $^{2}$Nanyang Technological University} \\
  {\normalsize $^{3}$Shanghai University of Sport \qquad $^{4}$Baidu} \\[3pt]
  {\normalsize Project page: \url{https://ren-ym.github.io/point-humanoid-tennis/}}
}

\begin{document}
\maketitle
\input{fig/teaser}
\begingroup
  \renewcommand{\thefootnote}{\fnsymbol{footnote}}
  \footnotetext[1]{Equal contribution.\quad\textsuperscript{\ensuremath{\dagger}}Corresponding author.}
\input{sec/0_abstract}    
\input{sec/1_intro}
\input{sec/2_related_work}
\input{sec/3_method}
\input{sec/4_Experiment}
\input{sec/5_conclusion}

\section*{Acknowledgments}
This work was supported by the National Natural Science Foundation of China
(Project No.~62595774), the Key Laboratory of Intelligent Perception and
Human-Machine Collaboration (ShanghaiTech University), Ministry of Education,
the Shanghai QiYuan Innovation Foundation, and the HPC Platform of ShanghaiTech
University. We thank Shanghai University of Sport for providing the experimental
venues, the humanoid robot, and the equipment used in our real-world tennis
experiments.
% \clearpage
\section*{LLM Usage Statement}
OpenAI Codex was used to assist with text editing, figure-design guidance, code refinement, and experimental design, including final consistency and layout checks. The authors made the final decisions on the research, presentation, and reported results and remain responsible for the submitted content.
{
    \small
    \bibliographystyle{ieeenat_fullname}
    \bibliography{main}
}

% WARNING: do not forget to delete the supplementary pages from your submission 
\input{sec/X_suppl}

\end{document}

%% file: preamble.tex
\usepackage{cuted} %< for strip environment

\usepackage{currfile} %< for \currfilebase macro

\usepackage{caption} %< for "\captionof" commands (teaser)

%% file: fig/teaser.tex
\begin{strip}
\centering
\vspace{-8ex}
\includegraphics[width=\linewidth]{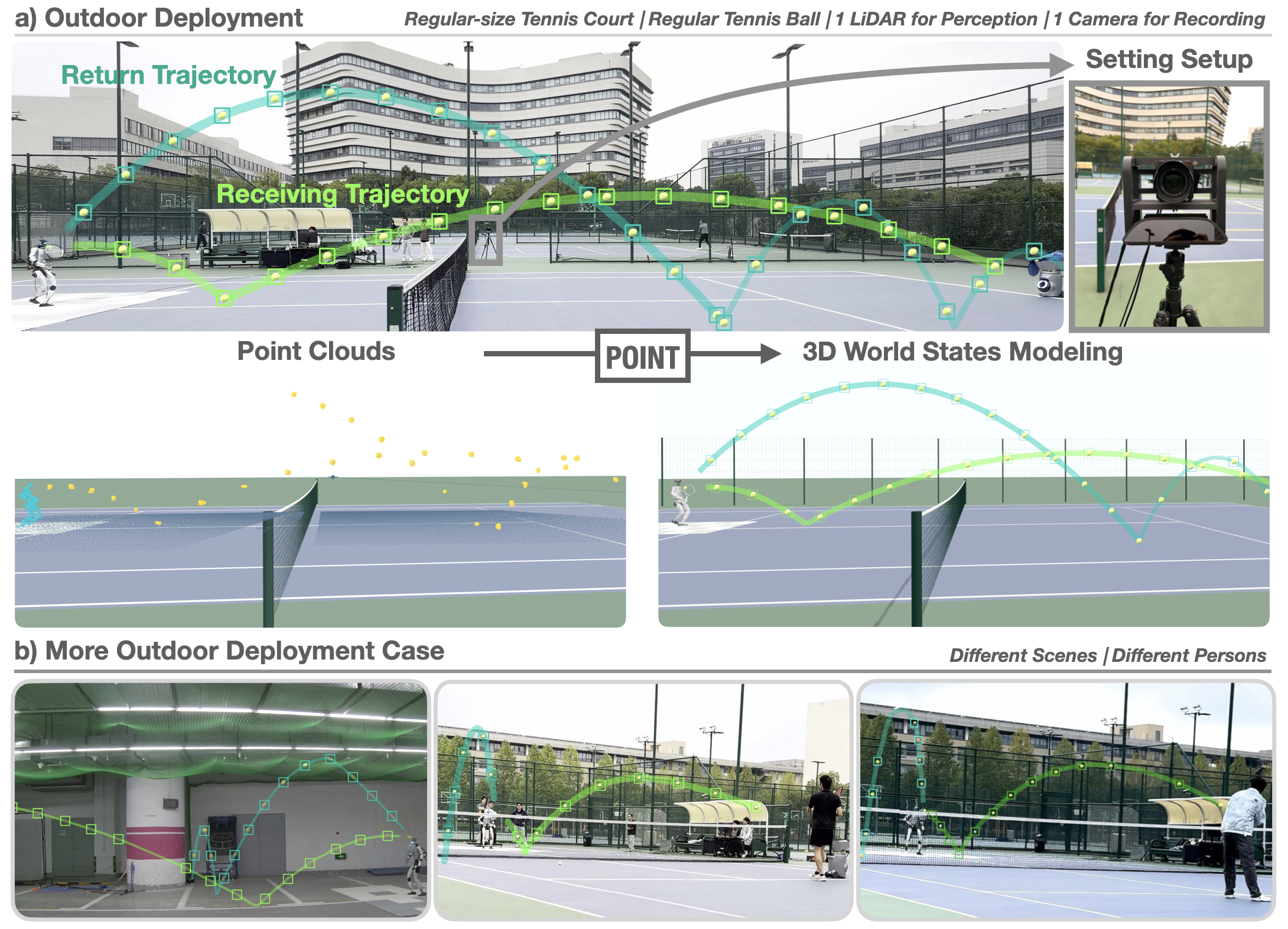}
\captionsetup{hypcap=false}
\captionof{figure}{
a) Real-world deployment on a regulation-size outdoor tennis court using a single external LiDAR and unmodified tennis balls. POINT reconstructs the humanoid and ball states online to support reactive interception and returns, with incoming and outgoing ball trajectories visualized alongside LiDAR observations and reconstructed 3D world states. b) Additional qualitative demonstrations of ball tracking across different indoor and outdoor environments and human players. The camera is used only for recording.}
\captionsetup{hypcap=true}
\vspace{-2ex}
\label{fig:teaser}
\end{strip}

%% file: sec/0_abstract.tex
\begin{abstract}

Humanoid tennis requires accurate 3D perception and coordinated whole-body control, yet existing systems rely heavily on motion-capture infrastructure, limiting their deployment on ordinary courts. We present \textbf{POINT}, a unified perception-to-action framework that enables reactive humanoid tennis using a single external LiDAR and onboard proprioception. To recover continuous 3D world states from sparse, asynchronous observations, POINT combines multimodal humanoid localization with physics-guided ball tracking, integrating proprioceptive motion, contact constraints, and bounce-aware trajectory refinement. To address uncertain interception decisions, we further introduce an Outcome-Aware Interception Planner that learns contact-time preferences and employs a rollout-supervised Contact-Event World Model to predict racket-ball interactions and return outcomes. This enables execution-aware plan selection from online observations without privileged future information. Under simulated perception, POINT achieves a $79.0\%$ task-return rate, compared with $53.9\%$ for our LATENT reproduction. On a regulation-size outdoor tennis court, POINT achieves a $70\%$ racket-contact rate and a $20\%$ successful-return rate across 20 trials with unmodified balls traveling at approximately $13.1$ m/s, despite restricted robot locomotion.

\end{abstract}

%% file: sec/1_intro.tex
\section{Introduction}
\label{sec:intro}
Enabling humanoid robots to interact autonomously beyond controlled laboratories is a central goal of embodied intelligence. Such autonomy relies on embodied perception to recover task-relevant 3D information about the robot and its surroundings. Maintaining reliable perception and control in large, dynamic environments remains challenging without extensive tracking infrastructure. Humanoid tennis is a demanding testbed: returning a fast-moving ball requires tracking its flight and bounce and coordinating whole-body movements within tight time constraints. Achieving this on ordinary outdoor courts with unmodified balls and a compact sensing setup would advance scalable embodied perception and autonomous interaction.

Despite recent progress, existing humanoid tennis systems remain difficult to deploy beyond instrumented environments. LATENT~\cite{rw_zhang2026latent}, ATHLETE~\cite{rw_tian2026athlete}, and AdaPT~\cite{rw_huang2026adapt} rely on external motion-capture infrastructure for their real-world rally demonstrations. LATENT employs over 50 cameras and reports approximately RMB 350,000 for three weeks of venue and infrastructure rental~\cite{rw_latent_deployment2026}, while ATHLETE operates within a $10.5\times5.5$ m tracked workspace. Similarly, AdaPT demonstrates indoor rallies on an $18\times6.5$ m court using 35 cameras and reflective tennis balls, while its vision-based alternative still requires six calibrated views. In contrast, CyboRacket~\cite{rw_ren2026cyboracket} avoids external motion capture through onboard RGB-D sensing, but demonstrates humanoid hitting only in a compact indoor environment, with limited evidence of reliable long-range outdoor perception. These limitations highlight the need for scalable 3D perception that supports dynamic humanoid interaction without extensive sensing infrastructure.

Bridging this gap requires continuously estimating the ball and humanoid states in a shared 3D frame. The ball is small and fast, with sparse observations and uncertain motion after bouncing, while humanoid localization is challenged by rapid movements, occlusion, and accumulated drift. Single-view LiDAR provides metric 3D measurements without distributed tracking infrastructure, but its 100 ms observation interval introduces substantial uncertainty in trajectory prediction and state estimation. More importantly, perception uncertainty directly affects interception planning: waiting for additional observations improves prediction reliability but reduces the time available for movement and racket preparation. Furthermore, different contact positions and times may lead to different execution outcomes, even along the same predicted trajectory. Reliable interception therefore requires evaluating multiple candidate contact plans based on both their predicted feasibility and execution outcomes, rather than relying on a single predicted contact point.

To address these challenges, we introduce \textbf{POINT}, as shown in Figure~\ref{fig:teaser}, a unified framework for online 3D world-state modeling and perception-conditioned humanoid tennis control. POINT continuously estimates the ball and humanoid states in a shared world frame using a single external LiDAR and onboard robot measurements. Our Multimodal Humanoid State Estimator fuses sparse LiDAR localization, high-frequency proprioception, and foot-contact constraints with acquisition-time alignment to limit drift. Our Physics-Guided Ball Tracker models aerodynamic flight and ground bounce, refining predictions as observations arrive. Together, they provide temporally consistent 3D world states for interception.
Building upon the motion prior of LATENT~\cite{rw_zhang2026latent}, we develop a Perception-Conditioned Policy and Planning framework that separates whole-body motion execution from contact planning. Through Teacher-Guided Policy Learning, a privileged teacher selects reference contact times to guide the training of a plan-conditioned policy under imperfect observations. However, teacher-selected contacts do not necessarily reflect the robot's actual execution capabilities. We therefore propose an Outcome-Aware Interception Planner (OAIP) that learns a contact-time preference distribution from teacher-guided executions and employs a Contact-Event World Model (CWM) to predict the physical outcomes of alternative plans. OAIP uses these predictions to select interception opportunities from online state estimates and adapt its decisions as observations arrive.

We compare POINT with LATENT and adapted ATHLETE under oracle ball states, simulated perception, and replayed real-world trajectories. Under simulated perception, POINT achieves a $79.0\%$ task-return rate, compared with $53.9\%$ for our LATENT reproduction. We further conduct ablation studies to evaluate the contributions of humanoid localization, physics-guided ball tracking, and outcome-aware interception planning. For real-world validation, we deploy POINT on a regulation-size outdoor tennis court ($23.77\times10.97$ m), testing long-range interception of unmodified balls traveling at approximately $13.1$ m/s. Despite uncontrolled outdoor conditions and restricted locomotion on protective mats, POINT achieves a $70\%$ racket-contact rate and a $20\%$ successful-return rate across 20 ball-launching trials using only a single external LiDAR and onboard sensors. Our main contributions are summarized as follows:
\begin{itemize}

    \item[$\bullet$] We demonstrate reactive humanoid tennis on a regulation-size outdoor court using \textbf{single-view LiDAR perception}, without distributed motion capture or modified tennis balls.

    \item[$\bullet$] We develop an \textbf{online 3D world-state modeling framework} that fuses asynchronous LiDAR, proprioception, and contact constraints with bounce-aware ball tracking, providing continuous estimates for interception under sparse observations.

    \item[$\bullet$] We propose an \textbf{Outcome-Aware Interception Planner} that couples learned contact-time preferences with a rollout-supervised Contact-Event World Model, separating promising contact proposals from execution-outcome evaluation.

    \item[$\bullet$] POINT returns balls more reliably than state-of-the-art humanoid tennis baselines when tracking incoming shots from sparse observations, both in simulated perception and in replay of real LiDAR recordings.

\end{itemize}

%% file: sec/2_related_work.tex
\section{Related Work}

\subsection{3D World Modeling for Embodied Interaction}

Reliable embodied interaction requires accurate 3D understanding of both the environment and the robot itself. Autonomous driving has advanced large-scale 3D perception and motion prediction using cameras and LiDAR~\cite{rw_bevformer2022,rw_centerpoint2021,rw_mtr2022,rw_uniad2023,rw_bevfusion2023,rw_pointpillars2019,rw_motionnet2020}, while robotic manipulation relies on object geometry and pose estimation for physical interaction~\cite{rw_foundationpose2024,rw_anygrasp2023,rw_densefusion2019}. For legged robots, state estimation combines inertial measurements, joint kinematics, and contact constraints to maintain continuous motion estimates~\cite{rw_bloesch2012,rw_hartley2020,rw_lin2022}. Pronto~\cite{rw_camurri2020pronto} further incorporates delayed, low-frequency exteroceptive measurements to correct high-rate proprioceptive estimates. While these methods provide foundations for embodied perception, dynamic racket sports require jointly estimating humanoid motion and fast-moving ball trajectories from sparse, asynchronous observations.

\subsection{Perception and Planning for Robot Sports}

Dynamic robot sports require tight coordination between perception, prediction, and physical control~\cite{rw_ma2025}. Robotic soccer combines visual perception with reactive locomotion~\cite{rw_tirumala2025,rw_wang2026soccer}, while table-tennis systems investigate high-speed ball tracking and interception~\cite{rw_gomez2019,rw_ziegler2025,rw_kreiser2026,rw_dambrosio2023robotic,rw_dambrosio2025competitive}. For humanoid table tennis, HITTER~\cite{rw_su2026hitter} integrates racket-target planning with reinforcement learning, while SMASH~\cite{rw_smash2026} demonstrates whole-body interaction using onboard vision. Humanoid badminton further explores trajectory prediction and dynamic striking~\cite{rw_liu2025badminton,rw_cui2026badminton}. Tennis introduces a larger interaction space, motivating court-scale sensing and trajectory prediction~\cite{rw_yang2023tennis,rw_zaidi2022esther}.

Recent humanoid tennis systems have advanced motion learning and interception planning, but remain constrained by their sensing configurations. LATENT~\cite{rw_zhang2026latent} learns athletic skills from imperfect motion references and demonstrates full-court rallies using extensive motion capture. ATHLETE~\cite{rw_tian2026athlete} combines trajectory-guided motion matching with teacher-student learning and reinforcement learning, while relying on external tracking for real-world evaluation. AdaPT~\cite{rw_huang2026adapt} develops adaptive motion planning and tracking for professional tennis styles, demonstrating indoor rallies with multi-camera motion capture or calibrated multi-view sensing. In contrast, CyboRacket~\cite{rw_ren2026cyboracket} integrates onboard RGB-D perception with physics-based trajectory prediction and whole-body control in a compact indoor environment. Despite these advances, robust interception under sparse, single-view observations remains challenging, particularly when uncertain ball predictions affect both contact selection and execution outcomes. POINT addresses this gap by combining online 3D world-state estimation with outcome-aware interception planning on a regulation-size outdoor tennis court.

%% file: sec/3_method.tex
\section{Methodology}
\label{sec:method}

\begin{figure*}[ht!]
    \centering
    \includegraphics[width=\linewidth]{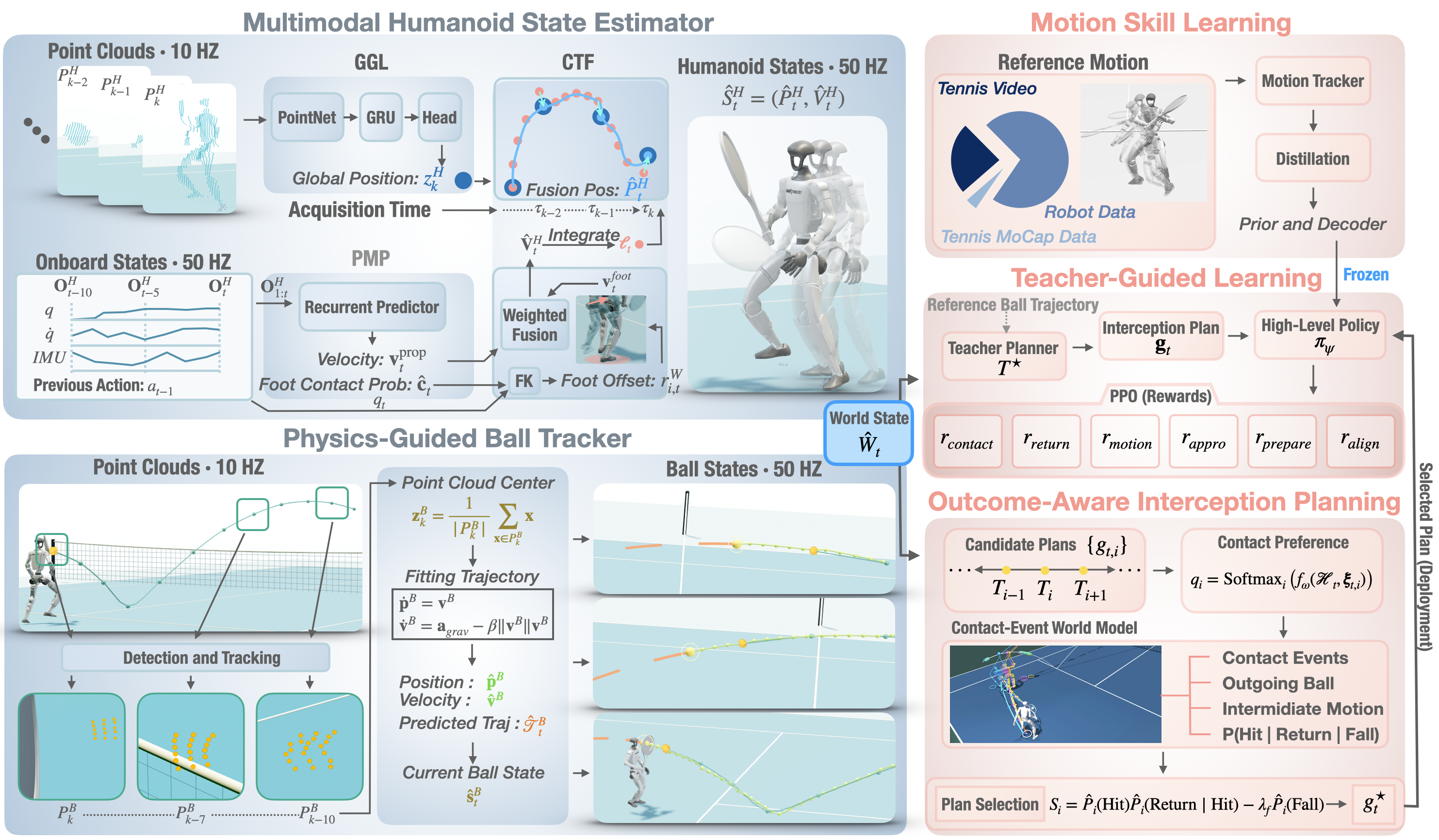}
    \caption{Overview of POINT. Online 3D world-state modeling estimates humanoid and ball states from single-view LiDAR and proprioception. Motion tracking and distillation provide reusable skills, while teacher-guided learning trains a plan-conditioned execution policy. OAIP selects interception plans using learned contact preferences and predicted execution outcomes.}
    \label{fig:pipeline}
    \vspace{-3ex}
\end{figure*}

POINT combines Online 3D World-State Modeling with Perception-Conditioned Policy and Planning (Figure~\ref{fig:pipeline}). The former estimates humanoid and ball motion. The latter pairs a plan-conditioned execution policy with OAIP, which selects plans using learned contact preferences and CWM predictions.

\subsection{Problem Formulation}
\label{sec:problem_formulation}

We use superscripts $H$ and $B$ for the humanoid and ball, respectively, and express global quantities in the court-aligned world frame $W$. Let $k$ index LiDAR scans and $t$ index control steps of duration $\Delta t$, with physical time $t\Delta t$. Each calibrated scan $\mathcal P_k=\{\mathbf x_{k,i}\}_{i=1}^{N_k}$ contains $N_k$ world-frame points $\mathbf x_{k,i}\in\mathbb R^3$ with acquisition timestamps $\tau_{k,i}$. The timestamp $\tau_k$ denotes the time associated with a scan-level measurement. Onboard measurements provide proprioceptive observations $\mathbf o_t^H$ and an IMU-aligned body-to-world rotation $\mathbf R_t^H\in\mathrm{SO}(3)$.
At each control step, POINT estimates the humanoid state $\hat{\mathbf s}_t^H=(\hat{\mathbf p}_t^H,\hat{\mathbf v}_t^H)$, ball state $\hat{\mathbf s}_t^B=(\hat{\mathbf p}_t^B,\hat{\mathbf v}_t^B)$, and future ball trajectory $\hat{\mathcal T}_t^B$, with positions and velocities in $\mathbb R^3$. Together with observation freshness and validity indicators $\boldsymbol\eta_t$, these form the world state $\hat{\mathcal W}_t=(\hat{\mathbf s}_t^H,\hat{\mathbf s}_t^B,\hat{\mathcal T}_t^B,\boldsymbol\eta_t)$. Hats denote estimates, while $\mathbf z$ denotes LiDAR-derived position measurements. We use $\hat{\mathbf p}_t^B(T)$ for the ball position predicted at future physical time $T$ using observations available by step $t$. All estimates are causal and timestamp-aware.

\subsection{Online 3D World-State Modeling}
\label{sec:world_state_modeling}

\subsubsection{Multimodal Humanoid State Estimator}
\label{sec:humanoid_state_estimation}

Our estimator combines low-frequency LiDAR localization (10 Hz) with high-frequency proprioceptive prediction (50 Hz) through geometric global localization (GGL), proprioceptive motion prediction (PMP), and contact-constrained temporal fusion (CTF).

\paragraph{Geometric Global Localization.}
Partial LiDAR observations make the humanoid point-cloud centroid sensitive to articulation and occlusion. Given the extracted robot points $\mathcal P_k^H$ and centroid $\bar{\mathbf p}_k^H$, we estimate the global base position measurement through a learned offset:
\begin{equation}
\mathbf z_k^H
=
\bar{\mathbf p}_k^H+
f_\theta(\mathcal P_k^H-\bar{\mathbf p}_k^H),
\label{eq:ggl}
\end{equation}
where $f_\theta$ combines PointNet~\cite{rw_pointnet2017} with a causal GRU to predict the offset from centered point clouds while retaining temporal context. The measurement $\mathbf z_k^H$ retains its acquisition time $\tau_k$. Following LiveHPS~\cite{rw_livehps2024}, we supervise the estimator using synthetic LiDAR scans of articulated robot meshes and ground-truth base positions.

\paragraph{Proprioceptive Motion Prediction.}
Between LiDAR updates, a recurrent predictor estimates body-frame base velocity and bilateral foot-contact probabilities from proprioceptive history and previous control commands:
\begin{equation}
\begin{aligned}
(\tilde{\mathbf v}_t^{\mathrm{body}},\hat{\mathbf c}_t)
&=f_\phi(\mathbf o_{1:t}^H,\mathbf a_{t-1}),\\
\mathbf v_t^{\mathrm{prop}}
&=\mathbf R_t^H
(\tilde{\mathbf v}_t^{\mathrm{body}}-\mathbf v_{\mathrm{bias}}),
\end{aligned}
\label{eq:pmp}
\end{equation}
where $\mathbf o_{1:t}^H$ is the observation history and $\mathbf a_{t-1}$ the previous action. The outputs are body-frame velocity and left/right contact probabilities. The stationary calibration bias $\mathbf v_{\mathrm{bias}}$ is removed before conversion to world-frame velocity $\mathbf v_t^{\mathrm{prop}}$. Training uses simulated base velocities, contact labels, and multi-step displacement supervision.

\paragraph{Contact-Constrained Temporal Fusion.}
We refine velocity using reliable support feet identified from contact probabilities and kinematic consistency. For foot $i$, its base-relative offset expressed in world coordinates is $\mathbf r_{i,t}^W=\mathbf R_t^H\operatorname{FK}_i(\mathbf q_t)$, where $\mathbf q_t$ denotes joint positions and $\operatorname{FK}_i$ computes the foot position relative to the base. Assuming stationary, non-slipping contact, the foot provides a base-velocity estimate $-\dot{\mathbf r}_{i,t}^W$. Aggregating valid foot estimates into $\mathbf v_t^{\mathrm{contact}}$, we obtain
\begin{equation}
\hat{\mathbf v}_t^H
=
(1-\lambda_t)\mathbf v_t^{\mathrm{prop}}
+\lambda_t\mathbf v_t^{\mathrm{contact}},
\label{eq:contact_fusion}
\end{equation}
where $\lambda_t\in[0,1]$ controls the contact contribution and is zero without reliable support.

We integrate the refined velocity into relative odometry $\boldsymbol\ell_t$ and maintain a global offset $\mathbf b_t$, initialized by the first valid LiDAR measurement. To correct drift without abrupt position changes, subsequent measurements are aligned with historical odometry at acquisition time $\tau_k$:
\begin{equation}
\begin{aligned}
\boldsymbol\ell_t
&=\boldsymbol\ell_{t-1}+\hat{\mathbf v}_{t-1}^H\Delta t,\\
\mathbf b_k^\star
&=\mathbf b_{t-1}+\mathbf K_k
[\mathbf z_k^H-\boldsymbol\ell(\tau_k)-\mathbf b_{t-1}],\\
\mathbf b_t
&=\mathbf b_{t-1}+\alpha_t(\mathbf b_k^\star-\mathbf b_{t-1}),\\
\hat{\mathbf p}_t^H
&=\boldsymbol\ell_t+\mathbf b_t.
\end{aligned}
\label{eq:global_fusion}
\end{equation}
Here, $\boldsymbol\ell(\tau_k)$ is interpolated odometry at acquisition time, $\mathbf K_k$ is the uncertainty-dependent measurement gain, and $\mathbf b_k^\star$ is the correction target from the latest valid LiDAR measurement. The release factor $\alpha_t\in[0,1]$ smoothly updates $\mathbf b_t$ at the control rate. Invalid measurements are rejected.

\subsubsection{Physics-Guided Ball Tracker}
\label{sec:ball_tracking}

Our tracker combines temporal association, aerodynamic trajectory fitting, and bounce-aware refinement to estimate ball states from sparse LiDAR observations.

\paragraph{Ball Observation and Trajectory Estimation.}
After filtering ground, background, and large objects, temporal association identifies ball clusters, yielding center observations $\mathbf z_k^B$ at acquisition time $\tau_k$. To avoid noisy finite differences, we fit observations to flight dynamics with gravity and quadratic air resistance:
\begin{equation}
\begin{aligned}
\dot{\mathbf p}^B &= \mathbf v^B,\\
\dot{\mathbf v}^B
&=\mathbf a_{\mathrm{grav}}-\beta\|\mathbf v^B\|_2\mathbf v^B,
\end{aligned}
\label{eq:ball_dynamics}
\end{equation}
where $\mathbf a_{\mathrm{grav}}$ is gravitational acceleration and $\beta$ is the calibrated drag coefficient. We estimate position and velocity at $\tau_k$ using observations from the same flight segment:
\begin{equation}
(\hat{\mathbf p}_{\tau_k}^B,\hat{\mathbf v}_{\tau_k}^B)
=
\arg\min_{\mathbf p,\mathbf v}
\sum_{j\in\mathcal I_k}
\rho\left(
\|\mathbf z_j^B-\Phi(\tau_j-\tau_k,\mathbf p,\mathbf v)\|_2^2
\right),
\label{eq:ball_trajectory_fitting}
\end{equation}
where $\mathcal I_k$ indexes recent valid observations, $\Phi(\delta,\mathbf p,\mathbf v)$ predicts position after interval $\delta$ under Eq.~\eqref{eq:ball_dynamics}, and $\rho$ is a robust penalty.

\paragraph{Bounce-Aware Online Prediction.}
We propagate the fitted trajectory to the predicted ground-contact time, obtaining pre-impact velocity $\mathbf v_b^-$. Given the court unit normal $\mathbf n$, the post-impact velocity is initialized as
\begin{equation}
\mathbf v_b^+
=
\alpha_{\parallel}(\mathbf I-\mathbf n\mathbf n^\top)\mathbf v_b^-
-\alpha_{\perp}\mathbf n\mathbf n^\top\mathbf v_b^-,
\label{eq:ball_bounce}
\end{equation}
where $\mathbf I$ is the $3\times3$ identity matrix, and $\alpha_{\parallel}$ and $\alpha_{\perp}$ denote tangential velocity retention and normal restitution. Subsequent post-bounce observations refine this prediction. The fitted state is propagated to the current control time, providing $(\hat{\mathbf s}_t^B,\hat{\mathcal T}_t^B)$ at 50 Hz. Valid partial-scan detections also allow early, causal updates of established tracks.

\subsection{Perception-Conditioned Policy and Planning}
\label{sec:interception_policy}

We separate plan-conditioned execution from selecting interception opportunities. Let $\mathcal H_t$ denote the causal history of estimated world states, proprioceptive observations, and previous actions. Our framework combines motion prior learning, teacher-guided execution training, and outcome-aware planning.

\paragraph{Motion Prior Learning.}
Following LATENT~\cite{rw_zhang2026latent}, we train a motion-tracking teacher and distill its skills into a latent motion model. Reference motions combine public human and robot datasets, professional tennis motion capture, and Internet tennis videos reconstructed with GVHMR~\cite{rw_gvhmr2024} and retargeted to the G1 using GMR~\cite{rw_gmr2025}. The teacher tracks these motions in physics simulation with randomized dynamics. Its actions are distilled into a reference-conditioned posterior, a proprioceptive motion prior, and an action decoder, providing reusable locomotion and racket skills.

\paragraph{Teacher-Guided Policy Learning.}
Under imperfect perception, a privileged teacher selects a reference contact time $T^\star$ using the estimated ball trajectory $\hat{\mathcal T}_t^B$, humanoid state $\hat{\mathbf s}_t^H$, and simulator reference trajectory:
\begin{equation}
T^\star=\arg\min_{T\in\mathcal C_t}
\left[
w_p e_t^2(T)+w_d d_t^2(T)
+w_v\left(\frac{d_t(T)}{T-t\Delta t}\right)^2
\right],
\label{eq:teacher_planning}
\end{equation}
where $\mathcal C_t$ contains valid future contact times, $e_t(T)=\|\hat{\mathbf p}_t^B(T)-\mathbf p_{\mathrm{ref}}^B(T)\|_2$ measures prediction error relative to the simulator reference position, and $d_t(T)$ is the required displacement to the target stance. The weights $w_p$, $w_d$, and $w_v$ balance prediction error, displacement, and required average speed.

Given $T^\star$, we construct an interception plan $\mathbf g_t$ from the \emph{online predicted trajectory}, containing robot-relative contact position and velocity, remaining time $T^\star-t\Delta t$, target stance, and observation validity. During teacher-guided training, the selected time remains fixed while new observations refine the contact target. The high-level policy $\pi_\psi(\mathcal H_t,\mathbf g_t)$ outputs latent and wrist actions, with the frozen decoder converting latent actions into joint commands. Following LATENT~\cite{rw_zhang2026latent}, we train $\pi_\psi$ using PPO~\cite{rw_ppo2017} with contact ($r_{\mathrm{contact}}$), return ($r_{\mathrm{return}}$), and motion-regularization ($r_{\mathrm{motion}}$) rewards, augmented by target approach ($r_{\mathrm{approach}}$), stable preparation ($r_{\mathrm{prepare}}$), and spatiotemporal alignment ($r_{\mathrm{align}}$). The motion prior and decoder remain frozen.

\paragraph{Outcome-Aware Interception Planning.}
At deployment, OAIP replaces privileged contact-time selection with learned contact preferences and CWM predictions. The preference model proposes promising opportunities, while CWM evaluates the execution outcomes of alternative plans.

Given $\hat{\mathcal T}_t^B$ and $\hat{\mathbf s}_t^H$, we sample candidate times $\{T_i\}$ and construct plans $\{\mathbf g_{t,i}\}$. Each feature vector $\boldsymbol\xi_{t,i}$ contains predicted contact position and velocity, remaining time $T_i-t\Delta t$, and target stance. A preference network $f_\omega$ ranks valid candidates:
\begin{equation}
q_i=
\frac{\exp(f_\omega(\mathcal H_t,\boldsymbol\xi_{t,i}))}
{\sum_{j\in\mathcal V_t}
\exp(f_\omega(\mathcal H_t,\boldsymbol\xi_{t,j}))},
\quad i\in\mathcal V_t,
\label{eq:contact_preference}
\end{equation}
where $\mathcal V_t$ indexes candidates satisfying perception and interception constraints. The normalized preferences $q_i$ are trained with Gaussian soft labels around successful, well-aligned teacher-selected times, supplemented by contact supervision from failed attempts. They represent relative preferences rather than calibrated return probabilities.

For each retained high-preference candidate, CWM predicts the consequences of executing its plan under the frozen policy. Conditioned on $\mathcal H_t$ and the exact candidate goal $\mathbf g_{t,i}$ provided to the execution policy, CWM predicts contact timing, pre-contact ball, humanoid, and racket states, and a joint distribution of outgoing ball position and velocity at racket--ball separation. Additional heads predict intermediate motion and probabilities of racket contact, net clearance, task return given contact, and falling. Training uses plan-conditioned simulation rollouts with recorded contact events, state transitions, and outcomes, combining masked event likelihood, intermediate-state regression, and outcome classification. Failed trials remain available for outcome supervision.

During execution, OAIP retains high-preference candidates and ranks them using CWM outcome predictions:
\begin{equation}
\begin{aligned}
S_i
&=\hat P_i(\mathrm{Hit})
\hat P_i(\mathrm{Return}\mid\mathrm{Hit})
-\lambda_f\hat P_i(\mathrm{Fall}),\\
\mathbf g_t^\star
&=\arg\max_{\mathbf g_{t,i}\in\mathcal G_t}S_i,
\end{aligned}
\label{eq:oaip_selection}
\end{equation}
where $\hat P_i(\mathrm{Hit})$, $\hat P_i(\mathrm{Return}\mid\mathrm{Hit})$, and $\hat P_i(\mathrm{Fall})$ denote predicted contact, conditional task-return, and fall probabilities, respectively. Here, $\lambda_f$ penalizes fall risk, and $\mathcal G_t$ contains the retained plans. The selected $\mathbf g_t^\star$ conditions the frozen execution policy, while temporal consistency constraints prevent unnecessary switching. OAIP re-evaluates plans as new LiDAR observations update $\hat{\mathcal W}_t$, without privileged future information.

%% file: sec/4_Experiment.tex
\begin{table*}[ht!]
\centering
\caption{Interception performance under three ball-observation settings. Hit, Net, and Return give the percentages of trials with racket contact, net clearance, and an in-court return, respectively. Bold indicates the highest rate within each difficulty group and overall.}
\label{tab:method_comparison}
\footnotesize
\setlength{\tabcolsep}{3.5pt}
\renewcommand{\arraystretch}{1.18}
\begin{tabular*}{\textwidth}{@{\extracolsep{\fill}}ll*{9}{c}@{}}
\toprule
& & \multicolumn{3}{c}{\textbf{Oracle Ball State}} & \multicolumn{3}{c}{\textbf{Simulated Perception}} & \multicolumn{3}{c}{\textbf{Real-Trajectory Replay}} \\
\cmidrule(lr){3-5}\cmidrule(lr){6-8}\cmidrule(l){9-11}
\textbf{Difficulty} & \textbf{Method} & Hit$\uparrow$ & Net$\uparrow$ & Return$\uparrow$ & Hit$\uparrow$ & Net$\uparrow$ & Return$\uparrow$ & Hit$\uparrow$ & Net$\uparrow$ & Return$\uparrow$ \\
\midrule
\multirow{3}{*}{\textbf{Easy}}
& LATENT~\cite{rw_zhang2026latent} & \textbf{99.27} & \textbf{91.41} & \textbf{89.01} & 93.80 & 71.53 & 65.43 & 90.91 & 55.00 & 20.00 \\
& ATHLETE (adapted)~\cite{rw_tian2026athlete} & 96.09 & 6.25 & 6.25 & 97.66 & 4.69 & 3.32 & 56.36 & 0.00 & 0.00 \\
& \textbf{POINT} & 98.05 & 84.52 & 80.62 & \textbf{97.80} & \textbf{86.52} & \textbf{81.64} & \textbf{96.82} & \textbf{74.09} & \textbf{47.73} \\
\midrule
\multirow{3}{*}{\textbf{Medium}}
& LATENT & 98.83 & \textbf{93.02} & \textbf{89.70} & 89.84 & 65.43 & 54.59 & 58.18 & 24.09 & 12.27 \\
& ATHLETE (adapted) & 92.19 & 8.59 & 7.03 & 88.87 & 8.20 & 4.69 & 42.27 & 1.36 & 1.36 \\
& \textbf{POINT} & \textbf{99.85} & 91.06 & 86.47 & \textbf{98.78} & \textbf{88.53} & \textbf{81.88} & \textbf{88.64} & \textbf{53.64} & \textbf{33.18} \\
\midrule
\multirow{3}{*}{\textbf{Hard}}
& LATENT & 92.53 & 78.76 & 74.71 & 78.61 & 49.80 & 41.65 & 26.36 & 4.09 & 3.64 \\
& ATHLETE (adapted) & 80.47 & 6.25 & 5.47 & 70.51 & 4.30 & 3.71 & 32.73 & 0.00 & 0.00 \\
& \textbf{POINT} & \textbf{94.58} & \textbf{82.08} & \textbf{77.69} & \textbf{91.75} & \textbf{81.05} & \textbf{73.34} & \textbf{50.00} & \textbf{12.27} & \textbf{8.64} \\
\midrule
\multirow{4}{*}{\textbf{Overall}}
& LATENT & 96.88 & \textbf{87.73} & \textbf{84.47} & 87.42 & 62.26 & 53.89 & 58.48 & 27.73 & 11.97 \\
& ATHLETE (released) & 2.34 & 0.00 & 0.00 & 1.25 & 0.00 & 0.00 & 3.18 & 0.00 & 0.00 \\
& ATHLETE (adapted) & 89.58 & 7.03 & 6.25 & 85.47 & 5.86 & 4.18 & 43.79 & 0.45 & 0.45 \\
& \textbf{POINT} & \textbf{97.49} & 85.89 & 81.59 & \textbf{96.11} & \textbf{85.37} & \textbf{78.96} & \textbf{78.48} & \textbf{46.67} & \textbf{29.85} \\
\bottomrule
\end{tabular*}
\end{table*}

\section{Experiments}
\label{sec:experiments}
We evaluate interception under three observation settings and in outdoor trials, with ablations of humanoid localization, ball tracking, and planning.

\subsection{Experimental Setup}
\label{sec:exp_setup}

\paragraph{Platform and evaluation metrics.}
We use a Unitree G1 humanoid equipped with a tennis racket. A single external LiDAR provides observations at 10 Hz, while onboard proprioception supports state estimation and control at 50 Hz. Our primary test site is an outdoor doubles tennis court ($23.77\times10.97$ m). Real-world trials deploy full POINT with OAIP's learned contact proposals and CWM selection, without external motion capture or modified tennis balls. Hit denotes racket--ball contact, Net requires the struck ball to clear the net, and Return additionally requires its first landing inside the opponent's court. All evaluation trials, including misses and failed returns, enter the denominator. In Table~\ref{tab:real_world_results}, Dist. is the launcher-to-robot distance, Speed is measured as the ball approaches the robot, and $N$ counts trials. We also evaluate humanoid localization and ball prediction errors.

\subsection{Comparison with Existing Methods}
\label{sec:exp_comparison}

We evaluate whether strong interception under ideal observations transfers to uncertain ball estimates. We compare POINT with our LATENT reproduction~\cite{rw_zhang2026latent} and two ATHLETE~\cite{rw_tian2026athlete} variants: the publicly released checkpoint and our adapted reproduction with POINT's return reward. ATHLETE targets shorter-range exchanges. The adapted version retains native control and physics, so its results are diagnostic references rather than matched-environment comparisons.

\textit{Oracle Ball State} provides simulator ball states. \textit{Simulated Perception} uses ball-center observations with noise, missed detections, and delays modeled from real LiDAR recordings. It evaluates tracking and control under observation uncertainty, without simulating the full point-cloud perception pipeline. \textit{Real-Trajectory Replay} feeds recorded ball-center observations and their timestamps to the online tracker. Recorded trajectories drive the incoming ball, while robot interactions and outgoing balls are simulated. Incoming balls span three difficulty levels according to speed, required displacement, and available interception time. Training and evaluation trajectories are disjoint.

Table~\ref{tab:method_comparison} shows that our LATENT reproduction achieves higher overall Return than POINT under oracle inputs, but degrades substantially under simulated perception and recorded trajectories. Uncertain ball estimates make contact timing important: early interception requires acting on uncertain predictions, while waiting leaves less time for execution. POINT considers both the estimated trajectory and predicted execution outcomes, yielding a 25.1 percentage-point Return gain under simulated perception and a clear replay advantage. This supports adapting interception to imperfect observations, although difficult recorded trajectories remain challenging.

The released ATHLETE checkpoint has low contact rates even with oracle states. Adaptation substantially improves contact, but successful returns remain limited. This reflects transfer difficulty in our setting rather than performance under ATHLETE's original protocol, and highlights the need to control outgoing trajectories.

\begin{table}[t]
\centering
\caption{Real-world tennis performance of POINT.}
\label{tab:real_world_results}
\footnotesize
\setlength{\tabcolsep}{2pt}
\renewcommand{\arraystretch}{1.15}
\begin{tabular*}{\columnwidth}{@{\extracolsep{\fill}}cccccc@{}}
\toprule
\textbf{Dist. (m)}
& \textbf{Speed (m/s)}
& \textbf{$N$}
& \textbf{Hit}$\uparrow$
& \textbf{Net}$\uparrow$
& \textbf{Return}$\uparrow$ \\
\midrule
20-22 & 13.4 & 10 & 60.0 & 30.0 & 30.0 \\
22-25 & 12.7 & 10 & 80.0 & 10.0 & 10.0 \\
\bottomrule
\end{tabular*}
\end{table}

\begin{figure}
    \centering
    \includegraphics[width=1\linewidth]{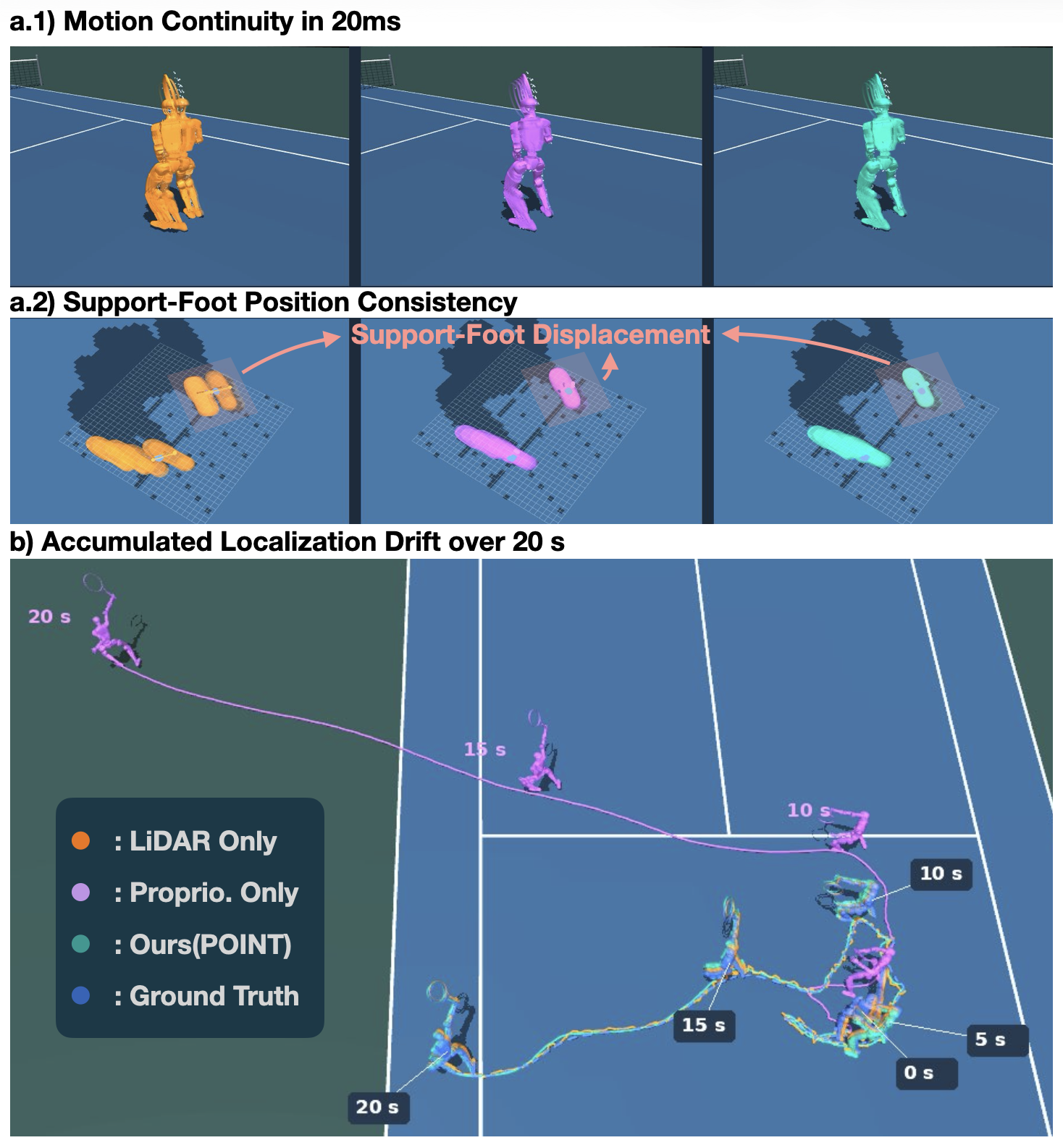}
    \caption{a) Motion continuity and support-foot consistency during high-frequency state updates. b) Accumulated localization drift over 20 seconds, comparing LiDAR-only localization, proprioceptive odometry, and POINT against ground truth.}
    \label{fig:humanoid_estimation}
\end{figure}

\begin{figure}[t!]
    \centering
    \includegraphics[width=1\linewidth]{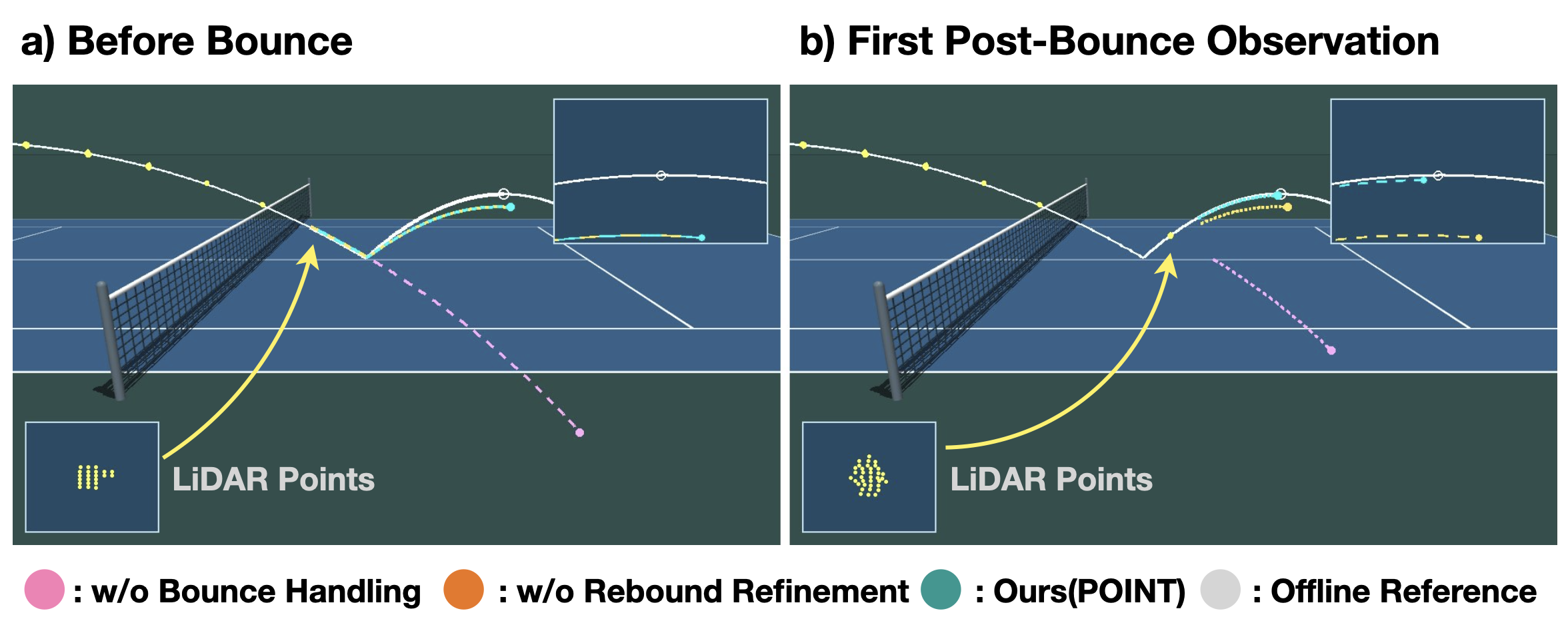}
    \caption{Predicted trajectories (a) before bouncing and (b) after the first post-bounce LiDAR observation. POINT progressively refines its prediction toward the offline reference, compared with variants without bounce handling or rebound refinement.}
    \vspace{-3ex}
    \label{fig:ball_tracking}
\end{figure}

\begin{table*}[t]
\begin{minipage}[t]{0.48\textwidth}
\centering
\caption{Humanoid state estimation over continuous 20\,s sequences without resets. $E_p$ and $E_v$ are mean Euclidean position and velocity errors, $D_{20}$ is accumulated displacement error, and $A_{95}$ is the estimated trajectory's 95th-percentile acceleration magnitude.}
\label{tab:humanoid_continuous_bias}
\footnotesize
\setlength{\tabcolsep}{3pt}
\begin{tabular*}{\columnwidth}{@{\extracolsep{\fill}}lcccc@{}}
\toprule
\textbf{Method} & $E_p\!\downarrow$ & $E_v\!\downarrow$ & $D_{20}\!\downarrow$ & $A_{95}\!\downarrow$ \\
 & (cm) & (m/s) & (cm) & (m/s$^2$) \\
\midrule
LiDAR only & \textbf{6.67} & 0.406 & 9.68 & 236.40 \\
Proprio. only & 297.01 & 0.543 & 848.19 & \textbf{10.42} \\
\textbf{POINT} & 6.91 & \textbf{0.241} & \textbf{8.21} & 17.02 \\
\bottomrule
\end{tabular*}
\end{minipage}\hfill
\begin{minipage}[t]{0.48\textwidth}
\centering
\caption{Mean Euclidean prediction errors (cm) at a 0.2-second horizon ($E_{0.2}$) and the first post-bounce update ($E_{\mathrm{first+}}$). Real-world evaluation uses offline fitted trajectories as pseudo-references.}
\label{tab:ball_ablation}
\footnotesize
\setlength{\tabcolsep}{2.5pt}
\renewcommand{\arraystretch}{1.10}
\begin{tabular*}{\linewidth}{@{\extracolsep{\fill}}l*{4}{c}@{}}
\toprule
& \multicolumn{2}{c}{\textbf{Simulation}} & \multicolumn{2}{c}{\textbf{Real}} \\
\cmidrule(lr){2-3}\cmidrule(l){4-5}
\textbf{Method} & $E_{0.2}\downarrow$ & $E_{\mathrm{first+}}\downarrow$ & $E_{0.2}\downarrow$ & $E_{\mathrm{first+}}\downarrow$ \\
\midrule
w/o bounce handling & 45.25 & 265.84 & 88.33 & 316.91 \\
w/o rebound refinement & 16.66 & 72.00 & 9.69 & 22.29 \\
\midrule
\textbf{Full POINT} & \textbf{10.59} & \textbf{29.70} & \textbf{7.85} & \textbf{17.36} \\
\bottomrule
\end{tabular*}
\end{minipage}
\par\medskip
\centering
\caption{Planning ablation. POINT variants share a frozen execution policy. Hit and Return are percentages. $E_{\mathrm{ball}}$ is mean Euclidean ball-position error (cm) at actual racket contacts, comparing online estimates with simulation ground truth or offline references at the same timestamp. Missed trials are excluded from this error metric. Bold denotes column-wise best results, including ties.}
\label{tab:planning_ablation}
\footnotesize
\setlength{\tabcolsep}{3.5pt}
\renewcommand{\arraystretch}{1.12}
\begin{tabular*}{\textwidth}{@{\extracolsep{\fill}}l*{9}{c}@{}}
\toprule
& \multicolumn{3}{c}{\textbf{Oracle Ball State}} & \multicolumn{3}{c}{\textbf{Simulated Perception}} & \multicolumn{3}{c}{\textbf{Real-Trajectory Replay}} \\
\cmidrule(lr){2-4}\cmidrule(lr){5-7}\cmidrule(l){8-10}
\textbf{Method} & Hit$\uparrow$ & Return$\uparrow$ & $E_{\mathrm{ball}}\downarrow$ & Hit$\uparrow$ & Return$\uparrow$ & $E_{\mathrm{ball}}\downarrow$ & Hit$\uparrow$ & Return$\uparrow$ & $E_{\mathrm{ball}}\downarrow$ \\
\midrule
LATENT & 96.88 & \textbf{84.47} & 0.00 & 87.42 & 53.89 & 5.64 & 58.48 & 11.97 & 5.39 \\
POINT + earliest & 77.88 & 36.78 & 0.00 & 81.48 & 45.74 & 6.96 & 61.36 & 10.30 & 5.52 \\
POINT w/o CWM & 96.61 & 80.27 & 0.00 & 95.05 & 77.16 & 4.86 & 76.97 & 28.48 & \textbf{4.50} \\
\textbf{POINT (full)} & 97.49 & 81.59 & 0.00 & 96.11 & 78.96 & \textbf{4.85} & \textbf{78.48} & 29.85 & 4.51 \\
POINT + GT/reference & \textbf{97.64} & 81.38 & 0.00 & \textbf{96.18} & \textbf{80.78} & 4.89 & 68.03 & \textbf{30.45} & \textbf{4.50} \\
\bottomrule
\end{tabular*}
\end{table*}

\subsection{Real-World Humanoid Tennis}
\label{sec:exp_real}

Table~\ref{tab:real_world_results} reports 20 outdoor ball-launching trials. Different incoming-ball distributions and physical conditions preclude a matched sim-to-real comparison of Return. Low-friction mats protecting the court restrict locomotion, while variable wind perturbs flight. These factors extend beyond the simulated observation model and may limit field performance.

Supplementary videos show qualitative deployments at 8--20\,m in underground parking garages and other indoor settings, and interception of human-hit balls on tennis courts.

\subsection{Ablation Studies}
\label{sec:exp_ablation}

We ablate state estimation, ball tracking, and planning while holding the remaining components fixed.

\paragraph{Humanoid State Estimation.}
We compare LiDAR-only localization, proprioceptive odometry, and POINT on continuous simulated sequences without resets.
Table~\ref{tab:humanoid_continuous_bias} shows the complementary limitations of the two sensing sources. Proprioceptive odometry produces smooth trajectories but accumulates substantial drift, while LiDAR-only estimates remain globally accurate but change abruptly between updates. POINT retains position accuracy close to LiDAR-only localization while reducing trajectory roughness by 92.8\%. Figure~\ref{fig:humanoid_estimation} illustrates how temporal fusion suppresses drift and maintains consistent support-foot positions between LiDAR updates.

\paragraph{Physics-Guided Ball Tracking.}
We ablate bounce handling and rebound refinement on matched observations at a 0.2\,s horizon, using simulation ground truth and offline fitted pseudo-references for real recordings.
Table~\ref{tab:ball_ablation} shows that explicitly modeling the bounce prevents the large errors caused by extrapolating flight through ground contact. Rebound refinement improves predictions in both settings, especially at the first post-bounce update, when timely estimates are critical for racket preparation. Figure~\ref{fig:ball_tracking} illustrates this correction toward the offline reference.

\paragraph{Interception Planning.}
All POINT variants in Table~\ref{tab:planning_ablation} share one frozen policy, incoming trajectories, candidate filters, and switching rules. The earliest variant selects the earliest valid contact candidate. Without CWM, the highest-probability contact time is selected. GT/reference uses ground truth or offline references only to select contact times. Execution goals still use online predictions. Earliest-point selection performs poorly even with oracle inputs, showing that a geometrically valid contact time need not be easy to execute.

The learned distribution recovers most of the improvement over earliest-point selection, indicating that successful interception depends strongly on choosing a favorable contact opportunity. CWM provides a smaller but consistent additional gain in Hit and Return across all three settings. On replay, it increases Return from 28.48\% to 29.85\%. Thus, outcome prediction refines the choice among promising contact opportunities. Full POINT approaches reference-guided Return on replay with higher Hit, so reference guidance is not a strict performance upper bound.

%% file: sec/5_conclusion.tex
\section{Conclusion}
\label{sec:conclusion}

We presented POINT, a perception-to-action framework enabling humanoid tennis on regulation-size outdoor courts using a single external LiDAR and onboard proprioception. By combining multimodal humanoid state estimation with physics-guided ball tracking, POINT maintains continuous 3D world states under sparse observations, while outcome-aware planning selects interception opportunities based on predicted execution outcomes. Experiments demonstrate the feasibility of this compact sensing setup. Real-world evaluation currently covers a limited number of trials with restricted locomotion. Reliable returns across broader conditions require further validation. Future work will improve racket control and locomotion and extend the framework to fully onboard perception.

%% file: sec/X_suppl.tex
% \clearpage
\appendix
\setcounter{secnumdepth}{2}
\setcounter{figure}{0}
\setcounter{table}{0}
\setcounter{equation}{0}
\renewcommand{\thefigure}{S\arabic{figure}}
\renewcommand{\thetable}{S\arabic{table}}
\renewcommand{\theequation}{S\arabic{equation}}
\renewcommand{\theHfigure}{supp.\arabic{figure}}
\renewcommand{\theHtable}{supp.\arabic{table}}
\renewcommand{\theHequation}{supp.\arabic{equation}}
\twocolumn[{\centering\Large\bfseries Supplementary Material\par\medskip}]

\noindent This supplement describes the evaluation protocols, state estimators, training procedures, and deployment of POINT. Sections~A--H refer to this supplement. Supplementary figures, tables, and equations carry the prefix S.

\section{Evaluation Protocols and Metrics}
\label{sec:protocol}

\subsection{Incoming-ball data and observation settings}
The three settings in Table~\ref{tab:method_comparison} of the main paper differ in how the incoming ball is observed. In all three, the simulator resolves robot motion, racket contact, and the outgoing flight.

\paragraph{Simulated trajectories.}
% Oracle Ball State and Simulated Perception share the trajectory bank used in base-policy training. 
Oracle Ball State and Simulated Perception use the same simulated trajectory-generation setup as execution-policy training. For Easy, Medium, and Hard, the generator caps launch speed at 12.4, 16.2, and 18.4\,m/s, respectively, and samples nominal flight times of 1.00--1.50, 0.85--1.45, and 0.70--1.35\,s. It samples target strike locations within 1.2\,m of the court centerline for Easy, and on either side at lateral distances of 0.25--1.8\,m for Medium and 0.35--3.2\,m for Hard. For the final stage of execution-policy training, we used 30,336 trajectories and reserved 384 for evaluation (128 per group). 

For POINT and LATENT, each evaluation run pairs every launch trajectory with four sensing seeds, yielding $384\times4=1{,}536$ trials. Under Simulated Perception, these seeds vary the synthetic observation stream for the same launch. Oracle Ball State retains the same trial indices and supplies the exact current ball state in each trial. Across four simulation resets, POINT and LATENT reuse the same trajectory--sensing-seed pairs, yielding $1{,}536\times4=6{,}144$ trials per method per setting. The four outer seeds preserve initial positions and velocities, resample the smoothing coefficients applied to delayed joint commands, and seed subsequent simulation steps. In ATHLETE's native simulator, each checkpoint uses the same 384 trajectories with 16 sensing seeds per trajectory under Simulated Perception. Oracle evaluation uses one deterministic rollout per trajectory.

\paragraph{Oracle Ball State.}
At each 50\,Hz control step, the planner reads the ball's current position and velocity directly from the simulator. It predicts future positions from this state to select a contact point.

\paragraph{Simulated Perception.}
We sample 3D ball centers from the simulated flight at 10\,Hz with a randomized scan phase. We perturb center coordinates and acquisition timestamps (standard deviations of 1\,cm and 2\,ms, clipped at 3\,cm and 4\,ms, respectively) and delay delivery by 5--15\,ms. At each control step, the tracker estimates position and velocity using only centers that have arrived. The underlying simulated flight provides the reference trajectory.

\paragraph{Real-Trajectory Replay.}
A single LiDAR records each incoming flight as a sequence of timestamped 3D ball centers. For execution-policy training, 55 recorded flights yielded 1,419 training examples through changes in court placement and bounce dynamics. For evaluation, we replayed 110 flights from sessions disjoint from those used for recorded training. On each side of the robot, we placed the post-bounce point nearest 1\,m in height at a lateral distance of 0.35, 1.35, or 2.35\,m from the robot centerline (Easy, Medium, and Hard, respectively). These placements preserve the recorded speed and bounce timing while varying the lateral reach required. Each method thus faces $110\times3\times2=660$ trials.

For each recording, we robustly fit a flight model with aerodynamic drag and a ground bounce to the timestamped centers over the full observed flight. This offline fit provides a continuous reference position between measurements and supplies the incoming path for replay. We apply the same court placement to the fitted path and the measured centers. The online tracker receives only the centers available by each control step.

\subsection{Metric definitions}
Hit, Net, and Return rates use all trials as the denominator. Table~\ref{tab:ball_ablation} of the main paper measures position error at a 0.2\,s forecast horizon. $E_{0.2}$ averages over queries valid for all trackers, while $E_{\mathrm{first}+}$ uses the first received update after the bounce at the same forecast horizon. Table~\ref{tab:planning_ablation} reports $E_{\mathrm{ball}}$, the mean Euclidean distance in centimeters between the estimated and reference ball positions at the instant of racket contact, averaged over hits with a valid reference.

\section{Online Humanoid State Estimation}
\label{sec:humanoid}

\subsection{Geometric global localization}
For each LiDAR scan, geometric global localization (GGL) samples 256 points from the segmented robot cloud and subtracts their centroid. A shared point encoder with widths $(3,128,256,512,1024)$ and max pooling produces a scan descriptor. We append the three spatial extents, point count, inter-scan interval, arrival delay, and a validity flag. A two-layer MLP of width 512 and a single-layer GRU of width 512 process the resulting sequence. A $512\!\rightarrow\!256\!\rightarrow\!3$ head predicts the offset from the centroid to the robot base.

Synthetic LiDAR scans of articulated robot meshes provide supervision: the target offset is the ground-truth base position minus the observed centroid. We train with a robust position loss and auxiliary yaw and uncertainty losses. Point dropout and coordinate noise vary the visible geometry during training.

\subsection{Proprioceptive motion prediction}
At 50\,Hz, the proprioceptive motion predictor (PMP) processes joint positions, velocities, estimated torques, IMU readings, the previous joint command, and timing and validity features. A $132\!\rightarrow\!256\!\rightarrow\!256$ MLP followed by a two-layer GRU of width 256 predicts body-frame base velocity and the contact probability of each foot. Before rotating the velocity into the court frame, we subtract a body-frame bias estimated while the robot is stationary.

Simulated base velocities and foot contacts supervise the two online outputs. To penalize accumulated drift, we also rotate the predicted velocities into the world frame using the ground-truth orientations, integrate them over 0.1, 0.3, 0.5, and 1.0\,s, and compare each displacement with the corresponding ground-truth base displacement.

\subsection{Contact-constrained temporal fusion}
A stationary foot provides a base-velocity estimate from the negative time derivative of its base-relative position expressed in the court frame. We accept a foot estimate when four conditions hold for at least 0.04\,s: contact probability exceeds 0.85, sole height is within 0.10\,m of the ground, foot angular speed is below 1.2\,rad/s, and the estimate agrees with the proprioceptive prediction to within 0.65\,m/s. If two accepted estimates differ by more than 0.25\,m/s, we discard both for that update. Otherwise, we weight the accepted estimates by contact confidence and angular speed and blend them with the proprioceptive prediction. With no accepted foot estimate, we use the proprioceptive prediction alone.

We integrate the resulting velocity to obtain relative odometry. When a LiDAR base-position measurement arrives, we interpolate the buffered odometry at the scan's acquisition time and use the position residual at that time to update the global offset. The correction is applied gradually to the current position estimate. The odometry buffer spans 0.8\,s, and measurements older than 0.3\,s are discarded.

\section{Physics-Guided Ball Tracking}
\label{sec:ball}

Equations~(5)--(7) of the main paper describe how we fit a flight segment and initialize the velocity after a bounce. Here we detail how the tracker uses acquisition and arrival times to incorporate sparse LiDAR measurements, and how it refines the rebound velocity as post-bounce observations arrive.

\paragraph{Measurement timing.}
A LiDAR center enters the fit only after it arrives. The tracker compares it with the predicted ball position at the center's acquisition time, then propagates the updated estimate to the current control time. If a scan yields no accepted center, the tracker continues to propagate the previous fit until a new measurement arrives.

\paragraph{Refinement after the first bounce.}
Let $(t_b,\mathbf p_b)$ be the predicted bounce time and position, and let $\mathbf v_0^+$ be the rebound velocity from Eq.~(7) of the main paper. Until a post-bounce center arrives, the tracker predicts the outgoing flight from this prior. With each new center, it holds $(t_b,\mathbf p_b)$ fixed for the update and refines the outgoing velocity by minimizing
\begin{equation}
\label{eq:point_rebound_refinement}
\begin{aligned}
\mathcal L(\mathbf v) ={}&
\sum_{j\in\mathcal J_+}
\frac{\|\Phi(\tau_j-t_b,\mathbf p_b,\mathbf v)-\mathbf z_j^B\|_2^2}
     {\sigma_p^2} \\
&+ \frac{\|\mathbf v-\mathbf v_0^+\|_2^2}{\sigma_v^2}.
\end{aligned}
\end{equation}
Here $\mathcal J_+$ contains recent, already received centers acquired after $t_b$, and $\Phi$ predicts position under the flight dynamics of the main paper. The residual scales are $\sigma_p=0.03\,\mathrm{m}$ and $\sigma_v=0.8\,\mathrm{m/s}$. The first center contributes immediately. Close to impact, its predicted position changes little with velocity, so the prior has greater influence. Later centers constrain the outgoing velocity over a longer interval.

\section{Policy Learning and Outcome-Aware Planning}
\label{sec:learning}
This section supplements main-paper Sec.~3.3 with the candidate grid, rollout supervision for the preference model and contact-event world model (CWM), and online switching criteria.

\subsection{Execution-policy training}
\label{sec:training}
During training, the teacher uses the reference ball trajectory to select a contact time. The actor tracks the online forecast for that fixed time as new observations arrive. It outputs a 32-dimensional latent action and three wrist commands that override the corresponding outputs of the frozen decoder.

The motion library contains 1,344 segments (600,097 frames): 22.26\% of frames come from tennis motion capture, 75.89\% from retargeted video, and 1.85\% from LATENT. Final training episodes use native exact histories (20\%), simulated perception (30\%), or recorded replay (50\%), for POINT. Our LATENT reproduction uses the same episode mixture, simulator, action interface, PPO training procedure, and task rewards. Its actor and critic receive no candidate-goal features, and inference uses neither the preference model nor CWM. Figure~\ref{fig:coverage} shows the court coverage of the simulated and recorded training trajectories.

\begingroup
\graphicspath{{fig/}{figures/}}
\begin{figure*}[t]
\centering
\includegraphics[width=\textwidth]{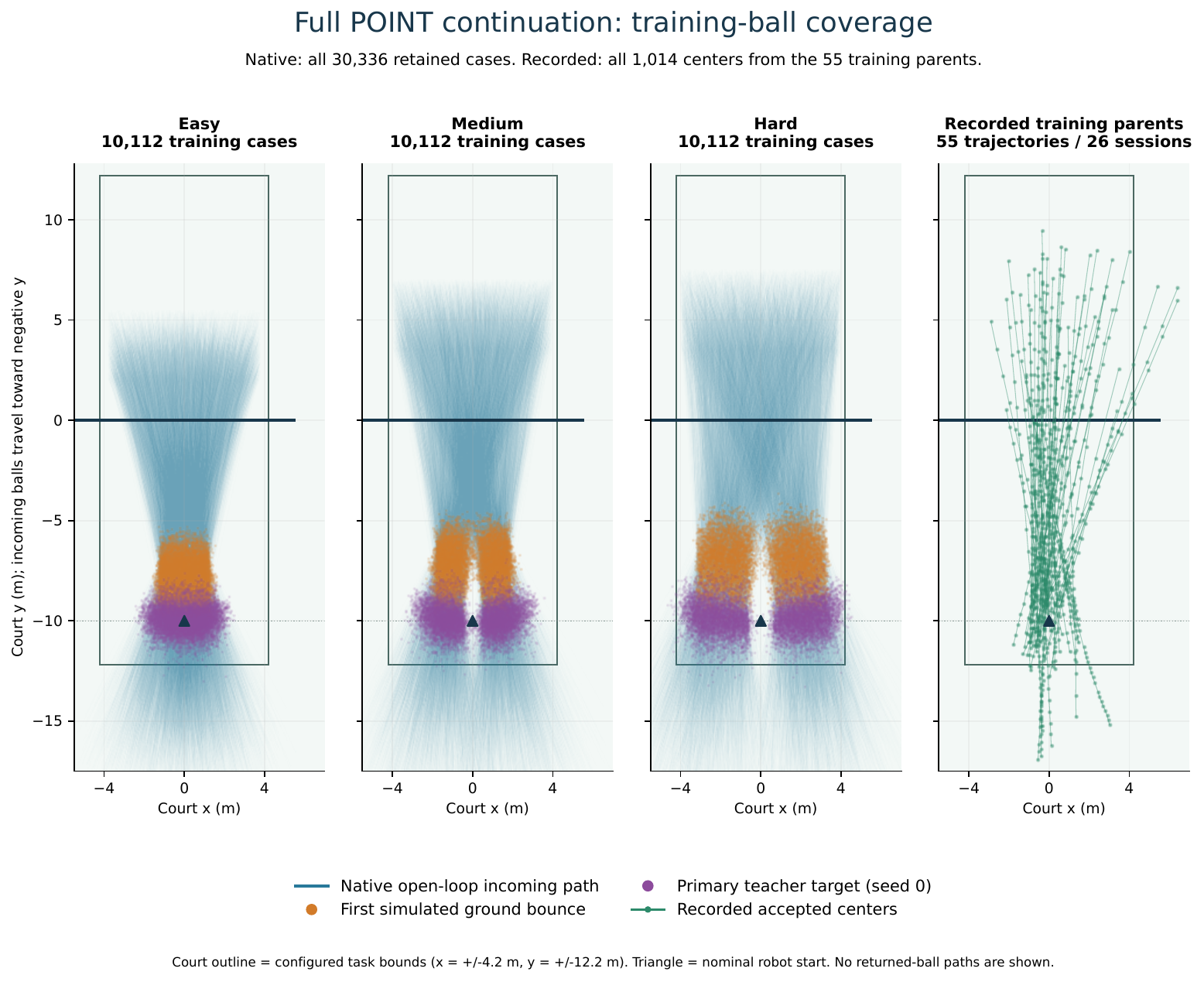}
\caption{\textbf{Spatial coverage of training incoming balls.} The first three panels show all 30,336 retained native cases, with 10,112 per difficulty. Blue curves are open-loop incoming trajectories, orange points are first ground bounces, purple points are privileged teacher targets for sensing seed 0. The last panel shows 1,014 accepted centers from 55 recorded training trajectories. Triangles indicate the nominal robot position. The outline denotes the configured simulation task bounds ($x=\pm4.2$ m, $y=\pm12.2$ m). %These are training inputs and targets, not executed robot paths or returned-ball trajectories.
}
\label{fig:coverage}
\end{figure*}

\endgroup

\paragraph{Execution rewards.}
Racket contact, net clearance, the first in-bounds landing, and terminal success have coefficients 20, 40, 120, and 300, respectively. The terminal penalties for a fall, miss, net failure, out-of-bounds landing, and second-bounce failure are 700, 240, 50, 50, and 260, respectively. Approach shaping rewards progress toward the target stance and readiness, with positive reward budgets of 25 and 10 per rally. Preparation shaping evaluates the predicted encounter near contact. At contact, alignment bonuses depend on stance error, the accuracy of the last pre-contact ball estimate, racket-face centering, and directional consistency. The final PPO stage uses a 20\,ms policy interval and adds approximately 403 million simulated frames.

\subsection{Candidate contact plans}
We sample 70 future contact times from 0.22 to 1.60\,s at 20\,ms intervals. A valid candidate requires a valid incoming estimate, finite predicted position and velocity, contact height in $[0.6,1.4]$\,m, and at least 40\,ms before the predicted second bounce. The current plan can replace the nearest grid time. Each candidate specifies robot-relative ball position and velocity, remaining time, target stance, and the displacement and average speed needed to reach it.

\subsection{Training the planning models}
A reference trajectory specifies where the ball will be at each time. Several points along that trajectory may admit a return, and the most favorable one depends on the robot's stance and the time available to prepare its swing. The teacher uses prediction error, stance displacement, and required speed to select a training target. 
% These criteria do not determine how well the execution policy will return the ball. 
We learn a preference over contact times from successful commands, then use CWM to compare the predicted outcomes of executing the most promising alternatives.

\paragraph{Contact-time preference.}
For each incoming case, the teacher supplies a reference time $T_{\rm ref}$. Five separate rollouts command $T=T_{\rm ref}+\delta$, with $\delta\in\{-160,-80,0,80,160\}$ ms. An in-court return supplies a preference label for that rollout at snapshots where the commanded time is admissible. Let $\mathcal N(T)$ contain valid grid times within 80 ms of $T$. With $\sigma_T=40$ ms, the target is
\begin{equation}
y_i=\frac{\mathbf{1}[i\in\mathcal N(T)]\exp[-(T_i-T)^2/(2\sigma_T^2)]}
{\sum_{j\in\mathcal N(T)}\exp[-(T_j-T)^2/(2\sigma_T^2)]}.
\label{eq:point_preference_target}
\end{equation}
Cross-entropy fits the preference distribution in main-paper Eq.~(9) to this target. Every tried admissible command also supervises the hit and return classifiers with its observed outcomes (binary cross-entropy weights 0.25 and 0.5). Failed returns contribute only to these classifiers.

Of 71,040 training rollouts, 43.24\% use oracle observations, 43.24\% simulated perception, and 13.51\% recorded replay. Each batch draws 64 examples per setting. % We transfer the preference model from an earlier execution-policy checkpoint.

\paragraph{History and network architecture.} Both planning models concatenate the current 24-dimensional ball packet and the two preceding packets with timing information. At the nominal 10 Hz cadence these span 0.2 s; packet ages preserve the actual timing. The preference model encodes an 87-dimensional context with an $87\!\to\!64\!\to\!64$ MLP, appends ten candidate features, and scores each candidate with a $74\!\to\!64\!\to\!32$ MLP and a linear head. The CWM state comprises 470 actor-context features, 13 base-state features, the three ball packets, the ages of the two previous packets, and the current time. The actor context includes five 50 Hz samples of joint position, joint velocity, base angular velocity, and projected gravity (0.08 s from first to last), plus four previous actions. A shared $558\!\to\!192\!\to\!192$ MLP encodes the state once. A $215\!\to\!192\!\to\!192$ MLP combines this encoding with 23 candidate features, including the exact 15 goal channels sent to the actor. Both models use SiLU activations.

\paragraph{Candidate commitment during CWM data collection.} At the branch decision, we fix the selected candidate’s absolute contact time for the remainder of that attempt. Until the final 0.20 s execution handoff, each eligible new ball observation updates the predicted contact position and incoming velocity at that same time. If the selected time becomes inadmissible, the update marks the plan invalid without switching to a different time. The frozen policy continues to act in closed loop, and the simulator records when contact actually occurs. Replanning across candidate times is enabled during deployment under the switching rules below.

\paragraph{CWM targets.}
Rollouts of the frozen policy pair each decision history and commanded plan with four targets:
\begin{itemize}
\item \emph{Motion.} An auxiliary head predicts 27 base and racket coordinates at a queried horizon up to 1.6\,s, supervised by mean squared error.
\item \emph{Contact.} A diagonal Gaussian models the difference between actual and commanded contact times and the ball, base, and racket states one physics step before impact (34 coordinates in total).
\item \emph{Separation.} A full-covariance Gaussian models the outgoing ball position and velocity at the first verified racket--ball separation.
\item \emph{Outcome.} The classifier uses the plan embedding and predicted event features. Its gradients stop at the event features.
\end{itemize}

\paragraph{Loss on hits and misses.}
Let $M$ contain the observed event coordinates and $A$ the available outcome labels for a trial. Using the training sampling weights, we minimize
\begin{equation}
\begin{aligned}
\mathcal L_{\mathrm{CWM}}={}&
 \mathbb E_w\!\left[\ell_{\mathrm{event}}(M)
 +0.5\,\ell_{\mathrm{out}}(A)\right]\\
 &+0.5\,\mathbb E_{w_h}\!\left[\ell_{\mathrm{motion}}(h)\right].
\end{aligned}
\label{eq:point_cwm_masked_loss}
\end{equation}
The event loss averages negative log-likelihood over observed coordinates. On a miss it is zero, and only hit and fall contribute to outcome binary cross-entropy. On a hit, the loss includes contact regression and conditional net and return classification. Separation regression requires verified separation. Motion supervision uses available future states on both hits and misses.

\paragraph{CWM validation.}
Training and validation use disjoint sets of 289 and 76 parent trajectories (79.2\% and 20.8\%), yielding 106,761 and 44,413 snapshots. We train with AdamW ($3\times10^{-4}$, batch size 256) and select epoch 36 on validation. 
% In Table~\ref{tab:point_cwm_dev}, timing and racket errors use valid contacts, outgoing velocity error uses verified separations, and the return Brier score uses all scored trials.
In Table~\ref{tab:point_cwm_dev}, timing and racket errors use valid contacts, outgoing velocity error uses verified separations, and the return Brier score is computed over trials with racket--ball contact.
\begin{table}[t]
\centering\small
\caption{CWM validation-set results by observation setting, balanced across trajectory sources.}
\label{tab:point_cwm_dev}
\begin{tabular}{@{}lrrr@{}}
\hline
Metric & Oracle & Simulated & Replay \\
\hline
Contact-time MAE (ms) & 13.63 & 21.35 & 26.85 \\
Racket position error (cm) & 9.19 & 11.06 & 13.17 \\
Outgoing velocity error (m/s) & 2.31 & 2.58 & 3.30 \\
Return Brier score & 0.114 & 0.129 & 0.127 \\
\hline
\end{tabular}
\end{table}

\subsection{Online selection and commitment}
Several contact times along the same incoming trajectory can lead to a successful return. The preference model proposes times supported by successful rollouts, and the CWM ranks those proposals by the predicted outcome of executing each command from the current robot state.
After a new ball observation, the preference model proposes up to three plans. CWM compares them with the current plan using the same decision state and main-paper Eq.~(10) ($\lambda_f=0.1$). A switch requires a utility gain above 0.03 after time and position switching costs, at least 100\,ms since the last switch, and more than 200\,ms until planned contact. An invalid plan can be replaced immediately. Otherwise, the policy follows the retained plan.

\section{Interpreting the Ablations}
The following comparisons clarify which information each variant retains in main-paper Tables~\ref{tab:humanoid_continuous_bias}--\ref{tab:planning_ablation}. Video times refer to the supplementary video.

\paragraph{Humanoid localization (Table~\ref{tab:humanoid_continuous_bias}).}
LiDAR-only localization estimates the base position from each accepted scan. Proprioception-only odometry integrates motion between updates without an external position correction. Full POINT propagates velocity using proprioception and reliable foot contacts, then corrects accumulated drift with LiDAR. On a continuous 20\,s sequence, $D_{20}$ measures error in the net displacement, whereas $A_{95}$ measures abrupt changes in the estimated trajectory through its 95th-percentile acceleration magnitude. A smooth estimate can therefore drift far from the reference. In the forward, sideways, and backward walking clips (3:43--4:05), compare the separation from the simulated reference over time and the position changes when a new scan arrives.

\paragraph{Ball tracking (Table~\ref{tab:ball_ablation}).}
All variants receive the same timestamped ball centers. Without bounce handling, the tracker extrapolates free flight through ground contact. Without rebound refinement, it predicts a bounce but disables the explicit fit of outgoing velocity to post-bounce centers. Full POINT updates that velocity as soon as the first post-bounce center arrives. In the video (3:19--3:40), compare pink (no bounce handling) with orange (rebound prior) as the ball reaches the ground, then orange with teal (full tracker) after the next observation. Both error columns use a 0.2\,s forecast horizon, the first-post-bounce column selects the update at which that forecast is made.

\paragraph{Interception planning (Table~\ref{tab:planning_ablation}).}
All POINT variants use the same execution policy and candidate filters. \emph{Earliest} chooses the first admissible time, \emph{Without CWM} uses the learned contact preference, and \emph{Full POINT} ranks the shortlisted plans by predicted execution outcome. \emph{GT/reference} uses the reference trajectory to select the time under the teacher's criteria. The actor still aims at the online forecast for that time. 
%The preference model already filters for times that have led to returns. 
CWM can improve the choice when shortlisted commands produce different outcomes from the current robot state. 
%Its scope is limited to this shortlist and the retained plan, all executed by the same policy. 
Table~\ref{tab:planning_ablation} reports incremental Return gains of 1.32, 1.80, and 1.37 percentage points for oracle, simulated perception, and replay, respectively.

The video illustrates two effects of contact-time selection. In a real-trajectory replay (4:09--4:23), reference-guided timing and preference-only selection both lead to misses. With CWM selection, the racket contacts the ball and the ball clears the net, although the trial does not yield an in-court return. The purple band contains the highest-ranked candidate times that together account for at least 80\% of the preference mass. In a separate oracle validation example (4:23--4:34), CWM scores three commands from the same decision state. All three produce contact when executed in simulation, but only the highest-scoring command yields an in-court return. The replay at 4:34--5:02 overlays the predicted and executed base and racket paths for this command and compares the ball states immediately before contact and at separation.

\paragraph{Returns under sequential feeding.}
We use 256 fixed sequences of up to 32 distinct incoming balls under Simulated Perception. Each in-court return allows another feed, with the robot state preserved between balls. The video (2:51--3:17) shows a longest sequence for each Table~\ref{tab:method_comparison} checkpoint:
12 returns for LATENT and 17 for POINT in the POINT simulator, and one for adapted ATHLETE in its native simulator, ATHLETE's native continuous-strike routine carries the robot pose, velocity, and previous action into the next cycle. 
% It specifies no separate target for returning to a ready position between feeds. In the sequence shown, ATHLETE returns the first ball in court and contacts the second ball, which lands short of the net.

\section{Physical Deployment and Supplementary Video}
\paragraph{From perception to contact.}
The opening sequence (0:20--0:41) connects the on-court setup to the point-cloud and simulation views. At 5:05--5:20, five physical contacts are shown with synchronized LiDAR, robot-pose, and camera views. The camera inset lets the viewer locate the stroke corresponding to each estimated pose. Online perception uses one external LiDAR and onboard joint and inertial measurements.

\paragraph{Selected deployment examples.}
The forehand and backhand clips (5:20--5:31) show each trial from two viewpoints. The next clips show lateral and forward approaches (5:31--5:46), a deep sideline interception (5:46--5:52), and feeds from different human players (5:52--6:03). The underground garage and indoor hall clips (6:03--6:15) show the system in two additional environments. The four-panel sequence at 6:15--7:09 shows all 20 outdoor trials from main-paper Table~\ref{tab:real_world_results}.

\paragraph{Video visualizations and AI use.} The policy-ablation replays overlay CWM predictions on recorded simulator executions. OpenAI image generation produced the two concept illustrations in the closing outlook (7:09--7:30). OpenAI Codex assisted with drafting and editing the English narration, and Kokoro text-to-speech synthesized the voice-over.

\section{Deployment Runtime}
\label{sec:runtime}

\paragraph{Deployment schedule.}
An external workstation (AMD Ryzen 7 9850X3D, NVIDIA RTX 5090 D v2) runs perception and control and communicates with the G1 over Ethernet. The RoboSense EM4 LiDAR supplies full scans at 10\,Hz. Proprioceptive prediction and the execution policy run at a nominal 50\,Hz, fused state estimates are published at 100\,Hz, and the host publishes joint commands at 500\,Hz. A policy target can therefore be written several times before the next policy update.

\paragraph{Physical deployment timing.}
Table~\ref{tab:runtime} reports host timings from a physical garage deployment of full POINT.
Frame processing starts after decoding. The fusion timer covers input handling, proprioceptive updates when due, and fusion through output construction. Geometric localization runs in a separate worker. The intervals overlap and have different sample counts.

\begin{table}[t]
\centering\small
\caption{Host timings in the physical garage session. P50 and P95 are the median and 95th percentile in milliseconds.}
\label{tab:runtime}
\begin{tabularx}{\columnwidth}{@{}>{\raggedright\arraybackslash}Xrrr@{}}
\toprule
Interval & $N$ & P50 & P95 \\
\midrule
Decoded-frame ROI processing & 5,012 & 6.89 & 7.62 \\
State-fusion loop work & 19,340 & 1.17 & 1.63 \\
Execution-policy inference & 503 & 0.27 & 0.44 \\
Policy query to first host write & 503 & 1.70 & 7.38 \\
Ball acquisition to first consuming write (estimated) & 124 & 31.08 & 129.04 \\
\bottomrule
\end{tabularx}
\end{table}

For each active policy query, we matched its target identifier to the first successful host DDS write carrying that target. Both endpoints use the host monotonic clock. To estimate first-use latency for a ball observation, we measured from its mapped LiDAR acquisition time to the first such write using that observation. All 503 queries had a matching write. Among them, 20 lacked a consistent acquisition timestamp pair. The remaining queries consumed 124 distinct observations, each counted once. The LiDAR-to-host clock mapping has an uncalibrated fixed transport offset. % Robot receipt and actuator response occur beyond the host write endpoint.

\paragraph{Offline planning benchmark.}
We timed OAIP separately on recorded ball packets and pre-action policy observations using one CPU thread on an Intel Xeon Gold 6330. Over 300 timed calls for one robot and three shortlisted candidates, feature assembly, trajectory forecasting, preference ranking, batched CWM evaluation, and final plan selection together took 11.58\,ms at P50 and 13.05\,ms at P95. The CWM step within that interval took 0.49 and 0.53\,ms, respectively. This replay interval begins with prepared inputs and ends with host plan construction.%  Live sensing, scheduling, and command transmission fall outside it.

\section{Limitations and Discussion}

\paragraph{Timing sensitivity on outdoor courts.}
The RoboSense EM4 supplies full scans at 10\,Hz, while the policy queries the predicted ball state at 50\,Hz. Between accepted ball-center observations, each 20\,ms query propagates the latest fit. At the incoming speeds measured near the robot (12.7--13.4\,m/s in main-paper Table~\ref{tab:real_world_results}), the ball travels about 13\,cm in 10\,ms. This scale illustrates why a small timing error near contact can shift where the ball meets the racket. An unmodeled crosswind could also shift the incoming path between observations. The outdoor trials did not quantify wind effects, so their contribution to the observed return rate remains unknown.

\paragraph{From individual returns to sustained rallies.}
To protect the court surface during outdoor trials, we used protective mats and restricted the robot's footwork. Sustaining a rally with a human partner on an ordinary court requires more consistent return placement and recovery after each stroke. A partner's pose before impact could help the robot prepare earlier, with subsequent ball observations refining the incoming trajectory estimate.

\paragraph{From courtside to onboard sensing.}
The external LiDAR requires placement and calibration at each court, and both players and the ball may move outside its field of view. Mounting the sensor on the robot would reduce courtside setup. Tracking from this moving viewpoint would require compensating for the robot's own motion and maintaining sight of the ball as the robot turns and swings.

\paragraph{Shared 3D state beyond tennis.}
POINT uses the estimated 3D states of the robot and ball to choose where and when to make contact. Extending this representation to include a human partner could support other interactions. In basketball, the passer's arm motion could provide an early estimate of the receiving region, which the robot could update from the ball's flight after release. In badminton, the opponent's stroke and a flight model specific to the shuttlecock could guide the timing and height of a return. In collaborative carrying, tracking both partners' hands and the object's pose could help the robot synchronize lift-off and keep the load level.